\documentclass[10pt,twocolumn,letterpaper]{article}

\usepackage{cvpr}              % To produce the CAMERA-READY version
\usepackage{graphicx}
\usepackage{amsmath}
\usepackage{amssymb}
\usepackage{booktabs}
\usepackage{multirow}
\usepackage[accsupp]{axessibility}

\usepackage[pagebackref,breaklinks,colorlinks]{hyperref}

\usepackage[capitalize]{cleveref}
\crefname{section}{Sec.}{Secs.}
\Crefname{section}{Section}{Sections}
\Crefname{table}{Table}{Tables}
\crefname{table}{Tab.}{Tabs.}

\def\cvprPaperID{10259} % *** Enter the CVPR Paper ID here
\def\confName{CVPR}
\def\confYear{2023}

\begin{document}

%%%%%%%%% TITLE - PLEASE UPDATE
\title{Perception and Semantic Aware Regularization for Sequential Confidence Calibration}

\author{Zhenghua Peng\textsuperscript{\rm 1}, Yu Luo\textsuperscript{\rm 1}, Tianshui Chen\textsuperscript{\rm 2}, Keke Xu\textsuperscript{\rm 1}, Shuangping Huang\textsuperscript{\rm 1,3,}\thanks{Corresponding author.}{ \ }\\
\textsuperscript{\rm 1}South China University of Technology, 
\textsuperscript{\rm 2}Guangdong University of Technology, 
\textsuperscript{\rm 3}Pazhou Laboratory\\
{\tt\small eepzh@mail.scut.edu.cn, luoyurl@126.com,  tianshuichen@gmail.com,}\\
{\tt\small eexkk@mail.scut.edu.cn, eehsp@scut.edu.cn}}
\maketitle

%%%%%%%%% ABSTRACT Pa zhou Laboratory
\begin{abstract}

Deep sequence recognition (DSR) models receive increasing attention due to their superior application to various applications. Most DSR models use merely the target sequences as supervision without considering other related sequences, leading to over-confidence in their predictions. The DSR models trained with label smoothing regularize labels by equally and independently smoothing each token, reallocating a small value to other tokens for mitigating overconfidence.
However, they do not consider tokens/sequences correlations that may provide more effective information to regularize training and thus lead to sub-optimal performance. In this work, we find tokens/sequences with high perception and semantic correlations with the target ones contain more correlated and effective information and thus facilitate more effective regularization. To this end, we propose a Perception and Semantic aware Sequence Regularization framework, which explore perceptively and semantically correlated tokens/sequences as regularization. Specifically, we introduce a semantic context-free recognition and a language model to acquire similar sequences with high perceptive similarities and semantic correlation, respectively. Moreover, over-confidence degree varies across samples according to their difficulties. Thus, we further design an adaptive calibration intensity module to compute a difficulty score for each samples to obtain finer-grained regularization. Extensive experiments on canonical sequence recognition tasks, including scene text and speech recognition, demonstrate that our method sets novel state-of-the-art results.
Code is available at \url{https://github.com/husterpzh/PSSR}.
%And the code will be released after review.  

\end{abstract}

%%%%%%%%% BODY TEXT
\section{Introduction}
\label{sec:intro}
Deep neural networks (DNNs) have shown remarkable performance in sequence recognition tasks, such as scene text recognition (STR)~\cite{FangXWM021,NguyenN0TNNH21,QiaoZYZ020} and speech recognition (SR)~\cite{KaritaSWDON19,BahdanauCSBB16}.
Despite impressive accuracy, recent studies have indicated that DNNs~\cite{chen2022knowledge,chen2022cross,huang2022agtgan, dai2023disentangling}, including deep sequence recognition (DSR) models, are usually poorly calibrated and tend to be overconfident~\cite{guo2017calibration,kumar2019calibration,kong2020calibrated}. In the sense that the confidence values associated with the predicted labels are higher than the true likelihood of the correctness of these labels, even for the wrong predictions, the overconfident DSR models may assign high confidences. This property may lead to potentially disastrous consequences for many safety-critical applications, such as autonomous driving~\cite{abs-1909-12358} and medical diagnosis~\cite{KimL20,mehrtash2020confidence}.

\begin{figure}
  \centering
  \begin{subfigure}{0.49\linewidth}
    \includegraphics[width=1\linewidth]{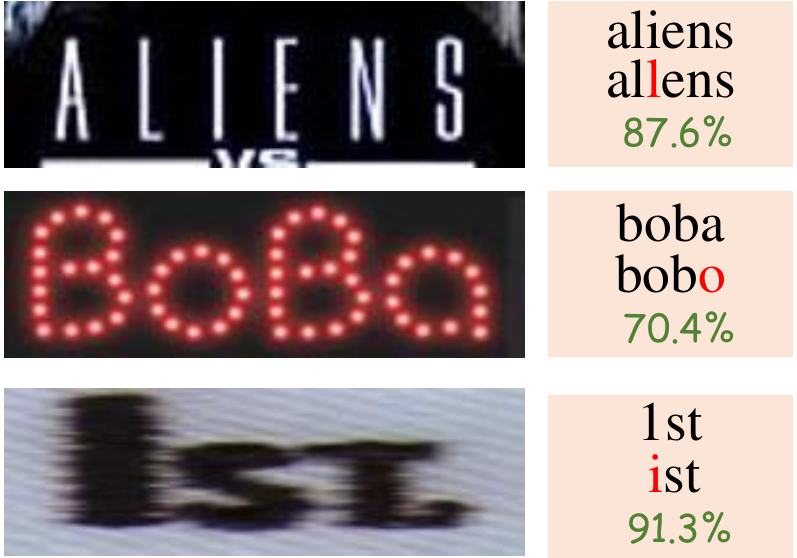}
    \caption{Perception overconfidence}
    \label{fig:sss-a}
  \end{subfigure}
  \hfill
  \begin{subfigure}{0.49\linewidth}
    \includegraphics[width=1\linewidth]{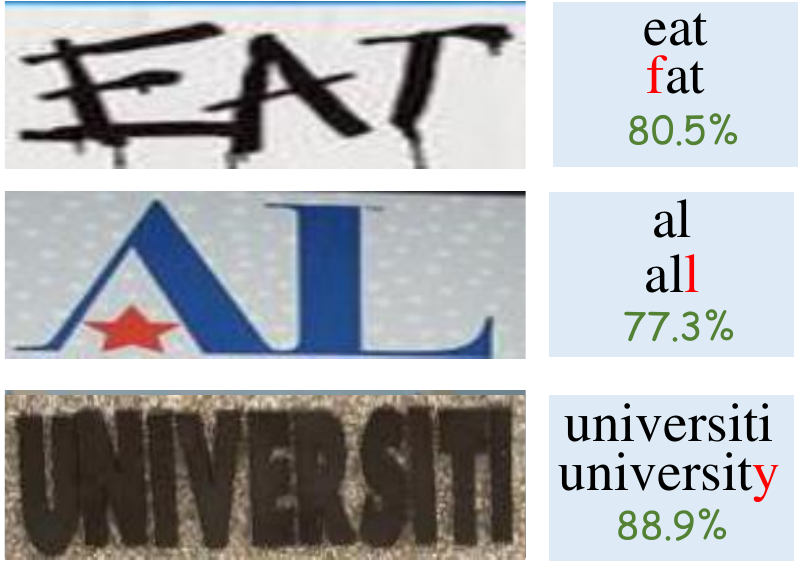}
    \caption{Semantic overconfidence}
    \label{fig:sss-b}
  \end{subfigure}
  \caption{Text strings placed along the right side of images are target, prediction, and sequence confidence respectively from top to bottom. \cref{fig:sss} \subref{fig:sss-a}: the model assigns higher confidence to the character that extremely resembles to the ground-truth character in visual perception (\eg, texture and topological shape); \cref{fig:sss} \subref{fig:sss-b}: the word that are semantically correlated to the ground-truth label will be predicted with a high confidence.}
  \label{fig:sss}
\end{figure}

Current DSR models use merely the target sequence as supervision and consider little information about any other sequences. Thus, they may tend to blindly give an overconfident score for their predictions, leading to the overconfidence dilemma. Presently, some works \cite{wang2020inference, GaoWHYN20} introduce label smoothing, which smooth each token by reallocating a small value to all non-target token class from the target class, to prevent the DSR models from assigning the full probability mass to target sequences.
However, these algorithms do not consider token/sequences correlations, and are difficult to provide effective and sufficient information. In this work, we find that tokens/sequences with high perception or semantic correlations, which refer to tokens/sequences with high visual/auditory similarities and with high co-occurrence similarities respectively, may be mistakenly given a highly-confident score. Taking STR for example, the Figure \ref{fig:sss} shows that token ``$l$" shares highly visual similarity with ``$i$", and thus the models may easily predict it to ``$i$" with high confidence. On the other hand, word ``universiti" is semantically similar to word ``university" and thus it is also predicted to ``university". These tokens/sequences are easily ambiguous with the target ones and thus may provide more effective information to regularize training.

In this paper, we propose a calibration method for DSR models:
\textbf{P}erception and \textbf{S}emantic aware \textbf{S}equence \textbf{R}egularization (PSSR).
The PSSR enables the DSR models with stronger vital perception discrimination ability and richer semantic contextual correlation knowledge by incorporating additional perception similarity and semantic correlation information into training.
Specifically, we construct a similar sequence set that comprises sequences either perception similar to the instantiated sequence input or semantic correlated with the target text sequence.
During the training stage, these similar sequences are used as weighted additional supervision signals to offer more perception similarity of different token classes and semantic correlation in the same context.
Furthermore, we discover that the degree of overconfidence of the model on its predictions varies across samples and is related to the hardness of recognizing samples. 
Hence, we further introduce a modulating factor function to adjust the calibration among different samples adaptively.  
To evaluate the effectiveness of the proposed method, we conduct experiments on two canonical sequence recognition tasks, including scene text recognition and speech recognition. 
Experimental results demonstrate that our method is superior to the state-of-the-art calibration methods across different benchmarks.

The major contributions of this paper are fourfold. 
First, we discovered the overconfidence of DSR models comprises perception overconfidence and semantic overconfidence.
Second, following our observations, we propose a calibration method for DSR models that enables the DSR models with more vital perception discrimination ability and richer semantic contextual correlation knowledge, so as to obtain more calibrated predictions.
Third, we introduce a modulating factor function to achieve adaptive calibration.
Fourth, we provide comprehensive experiments over multiple sequence recognition tasks with various network architectures, datasets, and in/out-domain settings. We also verify its effectiveness on the downstream application active learning. The results suggest that our method yields substantial improvements in DSR models calibration.
% The codes and trained models with be available for further research.

\section{Related Works}
\label{sec:related}

\textbf{Sequence Recognition.} 
Sequence recognition generally involves dealing with instantiated sequential data, which usually carries rich information on perception and semantic modalities. 
Previous methods, such as segmentation~\cite{RybachGSN09,DengLLC18} and CTC-based methods~\cite{ShiBY17,BahdanauCB14, GravesFGS06}, predict the sequence mainly depending on the perception feature of tokens, hardly taking semantic information into consideration.
For example, the CTC-based model splits the input sequence into several vertical pixel frames and outputs per-frame predictions, which are purely based on the perception features of the corresponding frame at each time step.
Recent works increasingly pay attention to conjointly exploiting both perception and semantic information~\cite{ZhangZYSL022,YuLZLHLD20, FangXWM021}, since the two types of information complement each other in the recognition process. Some works implicitly incorporate the semantic correlation to the models using RNNs with attention~\cite{ASTER,YangHZKG17,BahdanauCSBB16,00130SZ19} or Transformers~\cite{VaswaniSPUJGKP17,abs-2102-01547}.
Additionally, ~\cite{QiaoZYZ020,LuYQCGXB21,KaritaSWDON19} explicitly integrates a language model to learn semantics for supervision. 
Although remarkable progress has been achieved in the public benchmarks, we discover that it meanwhile incurs a problem, that is, these state-of-the-art methods are biased towards the commonly-seen perception pattern or the semantic context in the training set and produce overconfident predictions~\cite{WanZZLY20}.

\textbf{Confidence Calibration.}
Calibration of scalar classification has been extensively studied for a long time~\cite{2000Probabilistic,Niculescu-MizilC05,ZadroznyE01,2016Leveraging}. A simple yet efficient manner is post-hoc calibration, which directly rescales the prediction confidence of already trained models to the calibrated confidence during the inference stage~\cite{2000Probabilistic,guo2017calibration,JiJYKS19,patel2020multi}. While showing favorable effectiveness for in-domain samples calibration, they fail to be applied under the condition of dataset shift, since a held-out dataset is required to learn the re-calibration function~\cite{kong2020calibrated}.
As another prevalent line of research, several studies calibrate networks by modifying the training process during the training stage~\cite{brier1950verification}. 
Label smoothing~\cite{SzegedyVISW16}, originally proposed as a regularization technique, has shown a favorable effect on model calibration~\cite{pereyra2017regularizing}. 
\cite{pereyra2017regularizing} and \cite{mukhoti2020calibrating} fix overconfidence from the perspective of maximizing the entropy of the prediction distribution. 
More recently, Liu \etal argue that Label smoothing pushes all the logit distances to zero and lead to a non-informative solution, and propose a margin-based Label smoothing to realize better calibration~\cite{liu2022devil}. \cite{hebbalaguppe2022stitch} developed an auxiliary loss function that calibrates the entire confidence distribution in a multi-class setting.

The aforementioned methods mostly focus on the improvement and analysis of scalar classification task. However, almost little literature is proposed to study the calibration for DSR models calibration\cite{slossberg2020calibration,HuangLZYHW21}.
Slossberg \etal simply extend the temperature scaling for scene text recognition calibration, which rescale the logits on each time-step individually with a specific temperature value~\cite{slossberg2020calibration}. However, it essentially calibrates individual tokens to achieve the calibration of a sequence, unaware of perception and semantic correlation in calibration of sequences.
Huang \etal achieve the adaptive calibration on each token with taking the contextual dependency underlying the sequence~\cite{HuangLZYHW21}.  
Despite of showing a certain extent effectiveness, the method is insufficient for DSR models calibration, since only the inter-token context is considered, the potential cause of the overconfident prediction brought by the overfiting on the perception features are ignored.

%%%%%%%%%%%%%%%%%%%%%%%%%%%%%%%%%%%%%%%%%%%%%%%%%%%%%%%%%
% Please add the following required packages to your document preamble:
% \usepackage{multirow}
\begin{table*}[t]
\centering
\caption{The frequency ($F_{vis}$) and the average probability ($P_{vis}$) of ground-truth token being confused to most visually similar token}
\label{vis_pair}
\begin{tabular}{c|c|ccccccccccc}
\toprule
\multirow{3}{*}{CTC}  & Pair      & 0-$o$   & 1-$i$   & 3-$s$   & 4-$a$   & 5-$s$   & 8-$s$   & $c$-$e$   & $i$-$l$   & $l$-$i$   & $m$-$n$   & $o$-0   \\ \cline{2-13}
                      & $F_{vis}$      & 95.92 & 55.10 & 40.74 & 52.17 & 88.89 & 57.14 & 35.80 & 36.88 & 45.11 & 50.20 & 4.72  \\
                      & $P_{vis}$ & 84.01 & 75.02 & 60.92 & 78.79 & 76.87 & 76.57 & 68.10 & 70.86 & 69.11 & 66.23 & 65.13 \\ \midrule
\multirow{3}{*}{Attn} & Pair      & 0-$o$   & 3-2   & 3-5   & 5-$s$   & 8-$s$   & 9-$a$   & 9-$g$   & $c$-$e$   & $m$-$n$   & $n$-$m$   & $o$-0   \\ \cline{2-13}
                      & $F_{vis}$      & 70.00 & 50.00 & 50.00 & 72.73 & 44.44 & 40.00 & 40.00 & 37.33 & 40.75 & 56.64 & 2.58  \\
                      & $P_{vis}$ & 70.85 & 68.13 & 96.40 & 81.63 & 81.42 & 41.71 & 90.81 & 75.06 & 73.79 & 74.66 & 85.25 \\ \bottomrule
\end{tabular}
\end{table*}
%%%%%%%%%%%%%%%%%%%%%%%%%%%%%%%%%%%%%%%%%%%%%%%%%%%%%%%%%%%%%%%%%%%%%%%%%

\section{What Causes Overconfidence of DSR models?}
\label{sec3:cause}

In this section, we delve into reasons for the observed overconfidence of DSR models and identify that the perception similarity and semantic correlation of sequence are responsible for the phenomenon.
All the statistics are derived from the prediction results of a CTC-based model (NRNC~\cite{BaekKLPHYOL19}), and an attention-based model (MASTER~\cite{LuYQCGXB21}) on the ensembled testing set~\cite{JaderbergSVZ14,MishraAJ12,WangBB11,LucasPSTWYANOYMZOWJTWL05,KaratzasSUIBMMMAH13,KaratzasGNGBIMN15,PhanSTT13,RisnumawanSCT14}.

%%%%%%%%%%%%%%%%%%%%%%%%%%%%%%%%%
\begin{figure}[t]
\begin{center}
%\fbox{\rule{0pt}{2in} \rule{0.9\linewidth}{0pt}}
\includegraphics[width=1.0\linewidth]{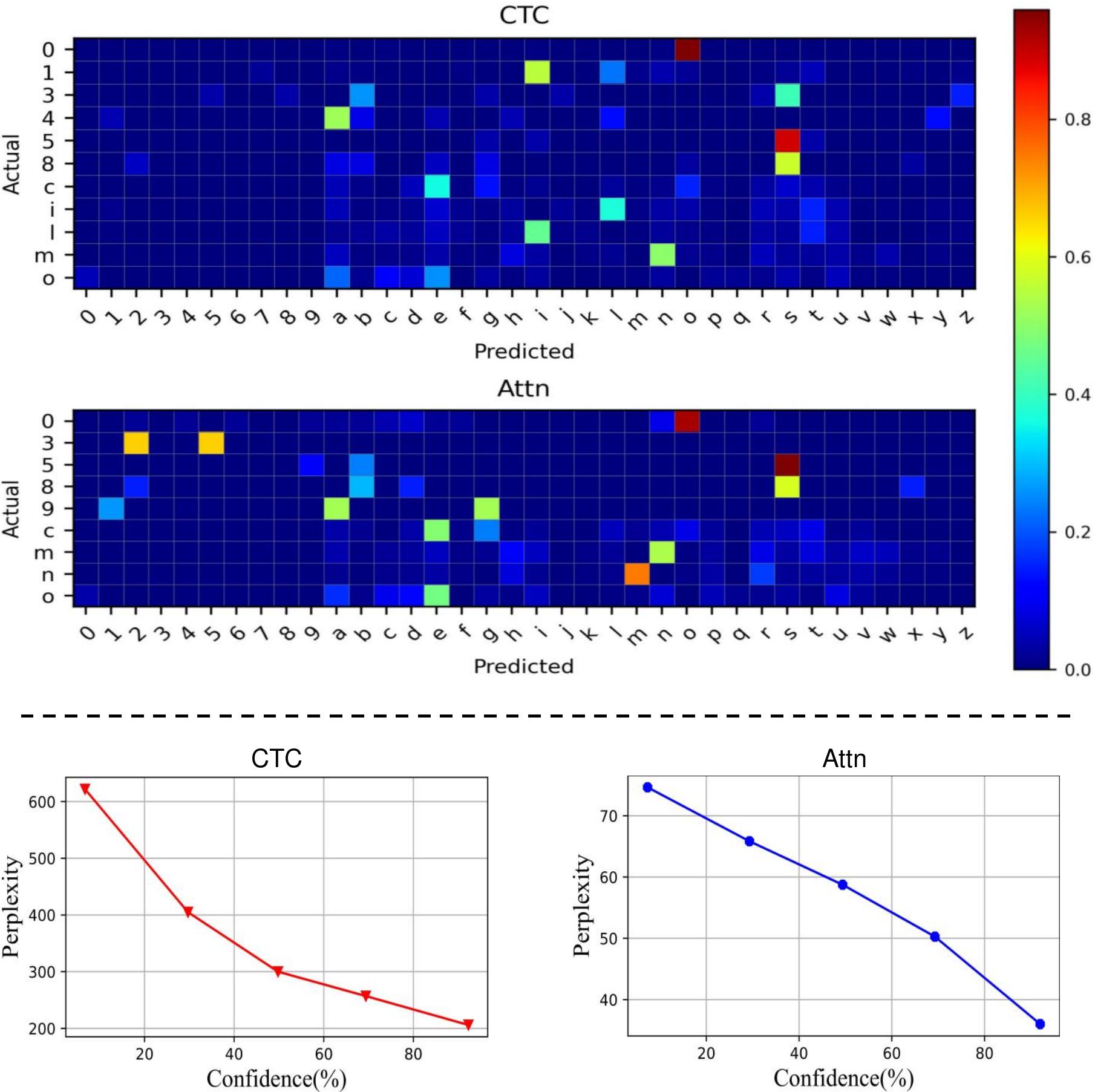}
\end{center}
  \caption{The upper part illustrates the confusion matrix of the mispredictions, which represents the distribution that the actual tokens in a sequence are recognized as other classes. And the bottom part plots the correlation between perplexity and the confidence of sequence.}
\label{v-s}
\end{figure}

\subsection{Perception Similarity}
To study how the perception information influences the miscalibration of DSR models, we build confusion matrices to count the frequency that the ground-truth class is recognized as other classes. 
The upper part of Fig.~\ref{v-s} displays part of the confusion matrices (see appendix for the complete confusion matrices), from which we can observe that the ground-truth class is more likely to be confused with the classes with higher perception similarity, that is, these classes suffer from more severe overconfidence. For example, the ground-truth token ``0’’ is almost exclusively confused with ``$o$’’ in either attention or CTC models.

Table~\ref{vis_pair} further presents the quantitative metrics, including the frequency ($F_{vis}$) and the average probability ($P_{vis}$) of the ground-truth token being confused to most perception similar token (see appendix for detailed calculation of the two metrics). As shown, a similar token owns a relatively high frequency and probability to be mispredicted. Note that, due to the data bias of the training set, where the proportion of letters is much larger than numbers, perception overconfidence mainly occurs in the letter-related classes. For example, the $F_{vis}$ in Table~\ref{vis_pair} show that, the letter ``$o$'' is seldom predicted as number ``0''.

\subsection{Semantic Correlation}

We additionally compute the perplexity of the misprediction of models to measure how well the predicted sequences are formed (see appendix for details of perplexity). In general, the lower perplexity score represents that the prediction has a stronger semantic correlation. As shown in the bottom part of Fig~\ref{v-s}, the semantically correlated mispredictions with lower perplexity scores demonstrate a more severe overconfidence problem.
Another interesting observation is that, although all the models tend overconfident in wrong predictions, the perplexity varies from the models based on different decoders. Compared with the CTC-based models that rely more on visual information, the context-aware attention-based models generally have lower perplexity scores. The phenomenon indicates that introducing the semantic information during training makes the model tend to predict legitimate sequences in the training set.

\section{Proposed Methodology}
\label{sec:method}

\subsection{Preliminaries}
\label{sec:method-0}
Let $\{(X_{i},Y_{i})\}_{i=1}^N \in \mathcal{D}(\mathcal{X}, \mathcal{Y})$ denotes a dataset, where $X_{i} \in \mathcal{X}$ is a sequential input sequence (\eg text image, speech audio, etc), and $Y_{i}=\{y_{i,1}, y_{i,2}, ..., y_{i,n_{i}}\} \in \mathcal{Y}$ is the corresponding target sequence consisting of multiple tokens. 
% For simplicity, we denote a sample as $(X,Y)$.
Let $\mathbb{P}(\Tilde{Y}|X_i)$ denotes the posterior probability that a sequence recognition network predicts for a candidate sequence $\Tilde{Y}$ on the given input $X_i$.
And the predicted sequence is obtained as $\hat{Y_i} = \mathrm{argmax}_{\Tilde{Y} \in \mathcal{Y}} \; \mathbb{P}(\Tilde{Y}|X_i)$ with its confidence as $\mathbb{P}(\hat{Y_i}|X_i) = \mathrm{max}_{\Tilde{Y} \in \mathcal{Y}} \; \mathbb{P}(\Tilde{Y}|X_i)$.
Generally, the DSR model are said to be perfectly calibrated when, for each sample $(X_i,Y_i) \in \mathcal{D}(\mathcal{X}, \mathcal{Y})$:
\begin{equation}
\label{cali}
\mathbb{P}(\hat{Y_i}=Y_i \mid \mathbb{P}(\hat{Y_i}|X_i)) = \mathbb{P}(\hat{Y_i}|X_i).
\end{equation}

\subsection{Sequence-level Calibration}
\label{sec:method-1}

The vanilla training process of the DSR model adopts one-hot encoding that places all the probability mass in one target sequence and thus encourages the probability of the target sequence being biased toward one-hot distribution.
This myopic training algorithm may be useful for recognition accuracy, but it ignores the perception similarity between different token classes and various semantic contextual correlations.
This lack of knowledge makes the model predict recklessly without considering various conditions.
To alleviate this problem, we attempt to incorporate additional information into the training stage, which comprises the perception similarity information between different token classes and more semantic contextual correlations.

Specifically, we construct a similarity sequence set that comprises sequences either perception similar to the sequence instance inside the input sequence or semantic correlated with the corresponding target sequence.
And we introduce a regularization term to the vanilla loss to smooth the empirical loss over these similar sequences.
Formally, the entire loss is defined as:
\begin{equation}
\label{eq:sssr}
    \mathcal{L}^{total}_i = \mathcal{L}_{G} (Y_i, \hat{Y_i}) + \alpha \; f(p_i) \sum_{Y^{\prime}_i \in \mathcal{S}(X_i,Y_i)} \mathcal{L}_{G} (Y^{\prime}_i, \hat{Y_i})
\end{equation}
where $\mathcal{L}_{G}$ refers to the empirical loss function (\eg, cross-entropy and CTC loss) generally used in DSR models of different decoding mechanisms, $\alpha$ is a hyperparameter used for adjusting the global calibration intensity, $f(p_i)$ is an adaptive calibration intensity function which is used for local adjustment of calibration intensity among different samples (see \cref{sec:method-3} for more details), and $\mathcal{S}(X_i,Y_i)$ is the similarity sequence set consisting of perception similarity and semantic correlation sequences of sample $(X_i,Y_i)$.

Most previous calibration methods for DSR models are implemented at the token-level, which require a token-to-token alignment relationship between input and output sequence and is therefore limited to the partial decoders (\eg, attention).
In contrast, our proposed loss is computed among different sequences, which can avoid the complicated alignment strategies operated on token-level and thus can be applied to different decoding schemes.

\subsection{Similar Sequence Mining }
\label{sec:method-2}
In this section, we describe how to obtain the similar sequence set, which consists of perception similarity and semantic correlation sequences

\textbf{Perception Similarity Sequences.}
The prediction distribution of DSR models is affected by both perception and semantic contextual features.
Thus, the critical challenge of effectively modeling the perception similar between sequences is to eliminate the effect of semantic context.
Hence, we resort to the semantic context-free model (\eg, CTC-based model). 
Specifically, we first fed the input sequences $X_i$ into a well-trained CRNN~\cite{ShiBY17} model to obtain the probability matrix consisting of  token prediction distribution at each time step.
Then, we can calculate the posterior probability $\mathbb{P}(\Tilde{Y}|X_i)$ of any candidate sequence $\Tilde{Y}$ over the entire sequence space.
Benefiting from the context-free attribute, the higher the probability of a candidate sequence, the higher its perception similarity to the input sequence.
Thus, we conduct a search algorithm based on the probability matrix to rank the posterior probability among the sequence space and finally collect the top \textit{N} probable sequences as the perception similarity sequences.

\textbf{Semantic Correlation Sequences.}
Recently, \cite{wan2020vocabulary} discovered models tend to assign high probabilities to sequences that share a highly similar context to the target sequence and appear more frequently in training, even if these sequences obviously deviate from the perception feature of the input sequence.
Here, we define them as semantic correlation sequences of the target sequence.
And we search for these sequences with the help of a pre-trained language model BCN~\cite{FangXWM021}, which is a variant of transformer decoder with a diagonal attention mask to prevent the model from attending to the current time-step token of the target sequence.
As a result, the token distribution at each time step is conditioned on its bidirectional context, that is $P(y_t|y_{1:{t-1}},y_{{t+1}:n})$. 
In this setting, we can efficiently model the correlation between tokens and their contexts.
Specifically, a higher probability for a certain token class means that the semantics of the combination of this token class with its context is stronger, i.e., this combination appears more frequently in training.
Then, we multiply the probability of each token together as the probability of semantic context correlation between the candidate sequence and the target sequence.
Similarly, we perform a search algorithm to rank the probability of sequences in the entire sequence space and collect the top \textit{N} probable sequences as the semantic correlation sequences of the target sequence.

\subsection{Hardness ranking adaptive calibration}
\label{sec:method-3}
\begin{figure}
  \centering
  \begin{subfigure}{0.49\linewidth}
    \includegraphics[width=1\linewidth]{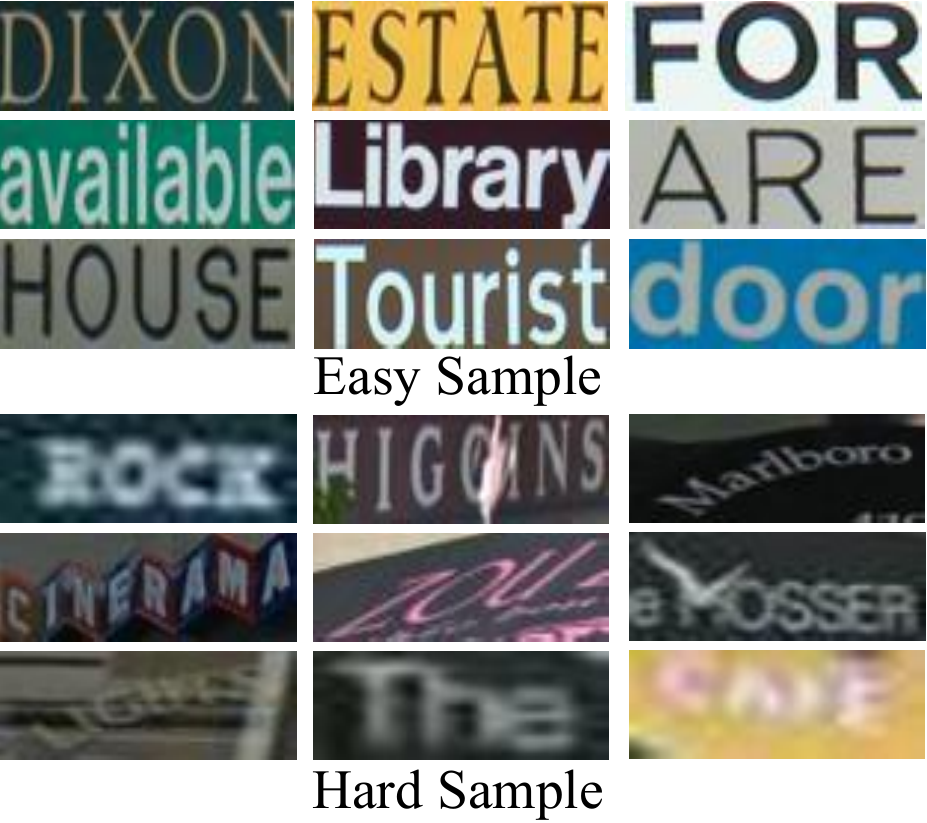}
    \caption{}
    \label{fig:hard-a}
  \end{subfigure}
  \hfill
  \begin{subfigure}{0.49\linewidth}
    \includegraphics[width=1\linewidth]{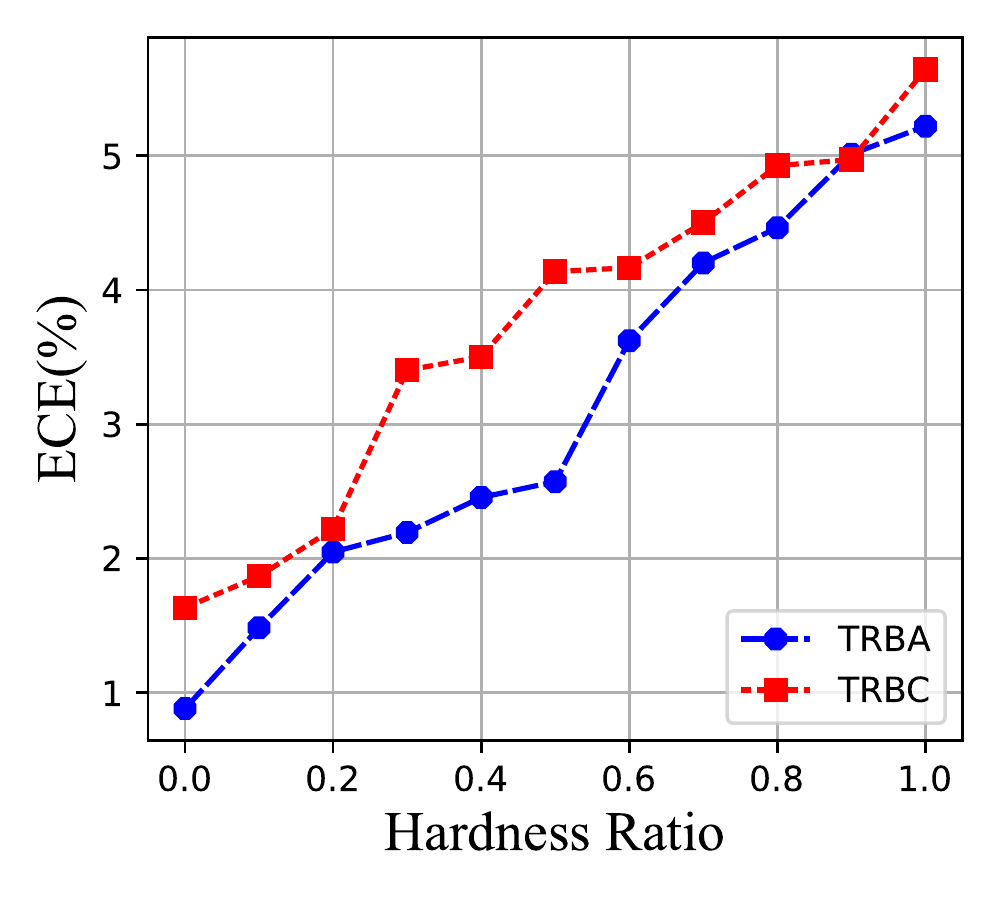}
    \caption{}
    \label{fig:hard-b}
  \end{subfigure}
  \caption{The illustration of: \textbf{(a)} easy and hard recognition samples; \textbf{(b)} the ECE results of TRBA and TRBC on different degrees of hardness of dataset.}
  \label{fig:hard}
\end{figure}

The models differ in the degree of overconfidence of their predictions on different samples, with more or less. Applying the identical calibration intensity to each sample may result in underconfident in some samples while may still be overconfident in others, which makes it challenging to achieve co-calibration.
To analyze the claim, we take STR as an example and construct a dataset with adjustable hardness property (see appendix for details).
We compare the calibration performance of TRBA~\cite{BaekKLPHYOL19} and TRBC~\cite{BaekKLPHYOL19} on the dataset with different hardness ratios.
As shown in \cref{fig:hard} \subref{fig:hard-b}, the ECE values all increase with increasing hardness ratio, indicating that the models become more overconfident.
One reason for this may be that the confidence of target sequences continuously increase when the model is trained with hard label, irrespective of the fact that the actual posterior probabilities of target sequences of difficult samples should be low intuitively. 
And the training process only leads to the predicted confidence scores become further greater than the actual probabilities.

Following our observation, and inspired by the \textit{Focal loss}~\cite{LinGGHD17} that views the posterior probability of the target class as a measure of the sample hardness (i.e. the lower the probability, the harder the sample), we propose a modulating factor function $f(p_i)$ that is integrated into the regularization term to achieve adaptive calibration based on sample hardness. 
It is defined as:
\begin{equation}
\label{eq:func}
    f(p_i) = \varepsilon_{e} + (\varepsilon_{h} - \varepsilon_{e})(1-p_i)^2
\end{equation}
where $\varepsilon_{e}$ and $\varepsilon_{h}$ are the hyper-parameters that control the calibration intensity for the easiest and hardest samples ($\varepsilon_{h} \geq \varepsilon_{e}$), respectively, and $p_i$ is the posterior probability of the target sequence (i.e. $\mathbb{P}(Y_i|X_i)$).
When the sample is hard to recognize and the $p_i$ is small, the result of $f(p_i)$ is close to the $\varepsilon_{h}$, so that more probability is smoothed from the target sequence towards similar sequences, and vice versa.
In this work, we set $\varepsilon_{e}$ and $\varepsilon_{h}$ are 0.01 and 1.0, respectively.

%%%%%%%%%%%%%%%%%%%%%%%%%%%%%%%ly%%%%%%%%%%%%%%%%%%%%%%%%%%%%%%%%%%%
\section{Experiment}

\subsection{Experimental Setup}

We evaluate our method on the two classic sequence recognition tasks: scene text recognition (STR) and speech recognition (SR). Detailed settings are described below.

\textbf{Datasets:}
For STR, we conduct the experiments on the English and Chinese benchmarks:
1) The English benchmark contains two synthetic datasets for training, i.e., Synth90K~\cite{JaderbergSVZ14} and SynthText~\cite{GuptaVZ16}, and the ensemble of seven realistic datasets for testing, including IIIT5K~\cite{MishraAJ12}, SVT~\cite{WangBB11}, IC03~\cite{LucasPSTWYANOYMZOWJTWL05}, 
IC13~\cite{KaratzasSUIBMMMAH13}, IC15~\cite{KaratzasGNGBIMN15}, SVTP~\cite{PhanSTT13}, and CUTE80~\cite{RisnumawanSCT14}. 
2) The Chinese benchmark~\cite{abs-2112-15093} ensembles five public datasets, consisting of 509,164 and 63,645 images for training and testing, respectively.
For SR, we use the AISHELL-1~\cite{8384449}, which is a large-scale mandarin speech dataset containing 141,600 sentences with 120,098 for training, 14326 for validation, and 7,176 for testing.

\textbf{Models:} 
For STR, we adopt six models, including ASTER~\cite{ASTER}, TRBA~\cite{BaekKLPHYOL19}, SEED~\cite{QiaoZYZ020}, MASTER~\cite{LuYQCGXB21}, CRNN~\cite{ShiBY17}, and TRBC ~\cite{BaekKLPHYOL19}, which cover the advanced and classical attention-based and CTC-based models. 
For SR, we use U2-Tfm\cite{abs-2012-05481} and U2-CTC\cite{abs-2012-05481}, which use a shared Comformer~\cite{peng2021conformer} encoder with self-attention and CTC as two branch decoders.

\textbf{Evaluation Metrics}: 
We adopt the widely used expected calibration error (ECE)~\cite{NaeiniCH15}, adaptive ECE (ACE) ~\cite{nguyen2015posterior}, maximum calibration error (MCE)~\cite{hebbalaguppe2022stitch}, and reliability diagram~\cite{2987588} as calibration metrics. Following~\cite{HuangLZYHW21}, these metrics are calculated by taking the entire sequence as a unit to calculate the sequence-level confidence and accuracy.

\textbf{Comparison Methods:} 
We compare our method with SOTA scalar and sequential calibration methods. Specifically, scalar calibration methods, including Brier Score (BS)~\cite{brier1950verification}, Label Smoothing (LS)~\cite{SzegedyVISW16}, Focal Loss (FL)~\cite{mukhoti2020calibrating}, Entropy Regularization (ER)~\cite{pereyra2017regularizing}, Margin-based Label Smoothing (MBLS)~\cite{liu2022devil}, and MDCA~\cite{hebbalaguppe2022stitch}, are extended to sequence recognition by applying them to each token. 
In addition, sequential calibration methods, including Graduated Label Smoothing (GLS)~\cite{wang2020inference}, Context-Aware Selective Label Smoothing (CASLS)~\cite{HuangLZYHW21}, are adopted for comparison. 
However, the two methods are limited to attention-based models due to the utilization of one-hot encoding.

\subsection{Ablation Study}
We conduct ablation studies to discuss and analyze the actual contribution of each component.
All the experiments are conducted on the English STR benchmark. TRBA and TRBC are adopted to validate the effectiveness of the proposed method on the sequence recognition models of attention and CTC decoders, respectively.

\begin{table}[t]
\caption{How hardness ranking adaptive calibration affects the sequence recognition calibration. The best method is highlighted in bold.}
\label{ablation:hr}
\resizebox{\linewidth}{!}{
\setlength{\tabcolsep}{1.2mm}{
\begin{tabular}{l|ccc|ccc}
\toprule
\multirow{2}{*}{Method}  &     & TRBA &     &     & TRBC &     \\  
                         & ECE & ACE  & MCE & ECE & ACE  & MCE \\ \midrule
PSSR w/o $f(p_i)$      & 0.74    &  0.93   &  8.97    &  1.19   & 0.85    &      10.65    \\ 
PSSR  &  \textbf{0.36}
   &  \textbf{0.28}
    &  \textbf{3.99}   &  \textbf{0.47}  &  \textbf{0.25}   &  \textbf{6.22}   \\ \bottomrule
\end{tabular}}}
\end{table}

\subsubsection{Effect of Adaptive Calibration}
As discussed in \cref{sec:method-3}, the hardness of recognizing a sample plays an important role in the calibration performance. 
We remove the component of hardness ranking $f(p_i)$ from the PSSR in Eq.~\ref{eq:sssr}, and show the comparison results of the PSSR with and without hardness ranking component in Table~\ref{ablation:hr}.
From the results, the resulting method suffers from a severe performance drop on all the metrics across TRBA and TRBC models. The calibration performance is more evident in the TRBA model, where the ECE, ACE, and MCE are increased by 0.38\%, 0.65\%, and 4.98\%, respectively.

\begin{figure}[t]
  \centering
  \begin{subfigure}{0.49\linewidth}
    \includegraphics[width=1\linewidth]{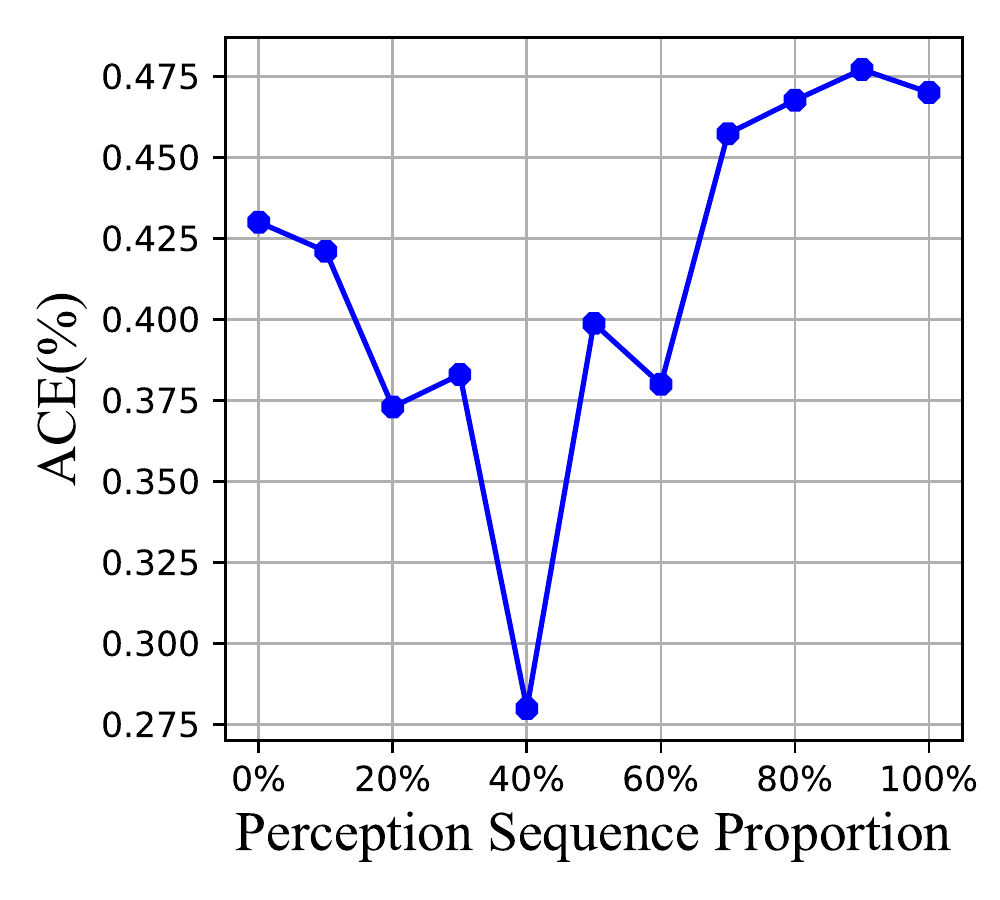}
    \caption{TRBA}
  \end{subfigure}
  \hfill
  \begin{subfigure}{0.49\linewidth}
    \includegraphics[width=1\linewidth]{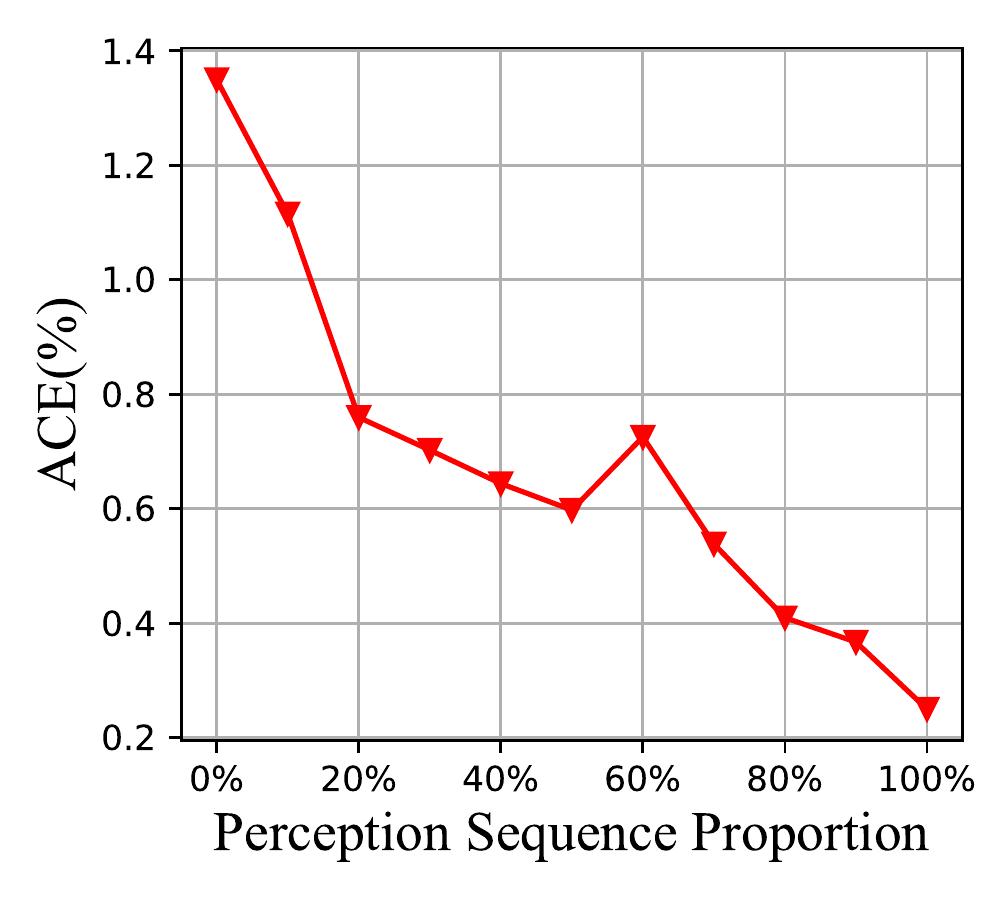}
    \caption{TRBC}
  \end{subfigure}
  \caption{The results of different perception similar sequence proportions in similarity sequence set on TRBA and TRBC models.}
  \label{proportion}
\end{figure}

\subsubsection{Effect of Similar Sequence Set}
The similar sequence set comprises both perception and semantic similar sequences. 
Here, we explore how the different combined proportions of these two kinds of sequences affect the calibration performance.
The results are shown in the \cref{proportion}.
As for TRBA, the calibration performance is better when the number of visually similar sequences is approximately equal to that of semantically similar sequences. However, the TRBC is better with the increase of visually similar sequences.
This confirms our claim above that the CTC-based models, such as TRBC, mainly occurring perception overconfidence, while overconfidence in attention-based models derives from both perception overconfidence and semantic overconfidence.

\begin{table*}[]
\centering
\caption{The calibration results comparison of NLL, BS, LS, FL, ER, MBLS, MDCA, GLS, CASLS and PSSR on the English STR benchmark of attention-based models. The accuracy and three calibration metrics: Acc(\%), ECE(\%), ACE(\%) and MCE(\%), are listed. The best method is highlighted in bold.}
\label{en-str}
\resizebox{\linewidth}{!}{
\setlength{\tabcolsep}{3.6pt}{
\begin{tabular}{l|cccc|cccc|cccc|cccc}
\toprule
\multirow{2}{*}{Method} & \multicolumn{4}{c|}{ASTER}                                 & \multicolumn{4}{c|}{TRBA}                                  & \multicolumn{4}{c|}{SEED}                                  & \multicolumn{4}{c}{MASTER}                                  \\ 
                        & Acc & ECE & ACE & MCE & Acc & ECE & ACE & MCE & Acc & ECE & ACE & MCE & Acc & ECE & ACE & MCE \\ \midrule
NLL        & \textbf{85.27}        & 3.82         & 3.82         & 17.10        & 85.51        & 3.88         & 3.88         & 21.49        & 85.34        & 4.04         & 4.04         & 23.09        & 84.52        & 3.86         & 3.86         & 16.01        \\
BS~\cite{brier1950verification}             & 85.17        & 3.46         & 3.41         & 16.53        & 86.06        & 3.44         & 3.42         & 23.72        & 85.20        & 4.14         & 4.14         & 21.23        & 85.83        & 3.26         & 3.26         & 16.17        \\
LS~\cite{SzegedyVISW16}             & 84.35        & 0.99         & 0.81         & 10.27        & 84.12        & 1.59         & 1.52         & 10.38        & 84.62        & 1.23         & 1.20         & 9.61         & 85.16        & 1.37         & 1.32         & 8.11         \\
FL~\cite{mukhoti2020calibrating}             & 84.94        & 1.79         & 1.40         & 9.55         & 85.34        & 1.36         & 0.99         & 11.04        & 85.89        & 2.23         & 2.24         & 16.01        & 84.86        & 1.22         & 0.97         & 7.37         \\
ER~\cite{pereyra2017regularizing}             & 76.33        & 7.25         & 7.21         & 23.85        & 85.64        & 1.31         & 1.10         & 9.18         & 85.50        & 1.07         & 0.95         & 13.73        & 85.09        & 1.40         & 1.02         & 10.86        \\
MBLS~\cite{liu2022devil}           & 84.42        & 1.12         & 1.03         & 7.63         & 84.51        & 1.34         & 1.16         & 9.47         & 84.55        & 1.39         & 1.38         & 10.22        & 85.01        & 1.03         & 1.05         & \textbf{5.72}         \\
MDCA~\cite{hebbalaguppe2022stitch}           & 85.09        & 2.18         & 2.14         & 10.70        & 85.98        & 1.50         & 1.44         & 7.85         & \textbf{86.08}        & 2.54         & 2.47         & 20.58        & 84.92        & 1.25         & 0.82         & 6.70         \\ 
GLS~\cite{wang2020inference}            & 84.12        & 0.93         & 0.71         & 6.36         & 83.83        & 0.92         & 0.90         & 7.17         & 85.13        & 1.26         & 1.11         & 11.24        & 85.05        & 2.66         & 2.64         & 11.76        \\
CASLS~\cite{HuangLZYHW21}          & 84.65        & 0.86         & 0.77         & 5.55         & 85.41        & 1.02         & 0.98         & 7.94         & 85.71        & 1.59         & 1.36         & 13.15        & 84.89        & 1.16         & 0.93         & 12.20        \\ 
PSSR           & 85.06        & \textbf{0.69}         & \textbf{0.48}         & \textbf{5.26}         & \textbf{86.45}        & \textbf{0.36}         & \textbf{0.28}         & \textbf{3.99}         & 85.54        & \textbf{0.94}         & \textbf{0.77}         & \textbf{7.48}         & \textbf{86.03}        & \textbf{0.78}         & \textbf{0.40}         & 8.36         \\ \bottomrule
\end{tabular}}}
\end{table*}

% Please add the following required packages to your document preamble:
% \usepackage{multirow}
\begin{table}[]
\caption{The calibration results of CTC-based models on the English STR benchmark. The best method is highlighted in bold.}
\label{en-str-ctc}
\resizebox{\linewidth}{!}{
\setlength{\tabcolsep}{0.85mm}{
\begin{tabular}{l|cccc|cccc}
\toprule
\multirow{2}{*}{Method} & \multicolumn{4}{c|}{CRNN}    & \multicolumn{4}{c}{TRBC}    \\ 
                        & Acc   & ECE  & ACE  & MCE   & Acc   & ECE  & ACE  & MCE   \\ \midrule
CTC                & 78.91 & 2.80 & 2.80 & 14.45 & 84.94 & 2.73 & 2.71 & 16.62 \\
PSSR                      & \textbf{79.53} & \textbf{0.97} & \textbf{0.49} & \textbf{8.52}  & \textbf{85.48} & \textbf{0.47} & \textbf{0.25} & \textbf{6.22}  \\ \bottomrule
\end{tabular}}}
\end{table}

% Please add the following required packages to your document preamble:
% \usepackage{multirow}
\begin{table*}[]
\centering
\caption{The calibration results comparison of NLL, BS, LS, FL, ER, MBLS, MDCA, GLS, CASLS and PSSR on the Chinese STR benchmark of attention-based models. The accuracy and three calibration metrics: Acc(\%), ECE(\%), ACE(\%) and MCE(\%), are listed. The best method is highlighted in bold.}
\label{zh-str}
\resizebox{\linewidth}{!}{
\setlength{\tabcolsep}{3.6pt}{
\begin{tabular}{l|cccc|cccc|cccc|cccc}
\toprule
\multirow{2}{*}{Method} & \multicolumn{4}{c|}{ASTER}   & \multicolumn{4}{c|}{TRBA}      & \multicolumn{4}{c|}{SEED}      & \multicolumn{4}{c}{MASTER}  \\ 
                        & Acc   & ECE  & ACE  & MCE   & Acc   & ECE   & ACE   & MCE   & Acc   & ECE   & ACE   & MCE   & Acc   & ECE  & ACE  & MCE   \\ \midrule
NLL        & 56.12 & 6.69 & 6.14 & 16.00 & 56.23 & 10.78 & 10.78 & 27.14 & 42.09 & 11.27 & 11.27 & 26.55 & 61.28 & 9.01 & 9.01 & 21.19 \\
BS~\cite{brier1950verification}                      & 56.18 & 6.14 & 5.58 & 16.03 & 56.82 & 10.18 & 10.18 & 25.34 & \textbf{44.15} & 10.41 & 10.41 & 24.55 & 65.06 & 8.77 & 8.77 & 20.88 \\
LS~\cite{SzegedyVISW16}                      & 56.31 & 1.95 & 1.53 & 4.71  & 56.16 & 1.25  & 1.23  & 4.18  & 42.54 & 1.31  & 1.34  & 4.21  & 65.28 & 1.33 & 1.33 & 3.22  \\
FL~\cite{mukhoti2020calibrating}                      & 55.98 & 5.73 & 5.19 & 12.95 & 56.78 & 9.74  & 9.74  & 24.58 & 43.02 & 8.69  & 8.72  & 21.45 & 64.04 & 2.96 & 2.96 & 7.43  \\
ER~\cite{pereyra2017regularizing}                      & 55.66 & 3.62 & 3.42 & 6.45  & 55.40 & 3.42  & 3.35  & 7.59  & 42.70 & 3.70  & 3.71  & 11.74 & 63.53 & 2.39 & 2.39 & 5.66  \\
MBLS~\cite{liu2022devil}                    & \textbf{56.32} & 1.96 & 1.52 & 5.15  & 56.26 & 1.29  & 1.18  & \textbf{2.29}  & 42.22 & 1.25  & 1.19  & \textbf{2.96}  & 65.66 & 1.13 & 1.13 & 3.31  \\
MDCA~\cite{hebbalaguppe2022stitch}                    & 56.06 & 5.63 & 5.08 & 13.01 & \textbf{56.85} & 9.97  & 9.97  & 26.74 & 43.43 & 9.68  & 9.69  & 22.78 & 64.12 & 3.02 & 3.02 & 8.93  \\ 
GLS~\cite{wang2020inference}                     & 56.16 & 1.38 & 1.15 & \textbf{2.83}  & 56.38 & 1.31  & 1.27  & 3.35  & 41.54 & 1.16  & 1.18  & 3.62  & 64.89 & 1.54 & 1.46 & 4.82  \\
CASLS~\cite{HuangLZYHW21}                   & 56.10 & 1.40 & 1.05 & 2.96  & 56.18 & 1.40  & 1.40  & 3.41  & 41.45 & 1.27  & 1.15  & 3.34  & 64.78 & 1.50 & 1.42 & 4.46  \\ 
PSSR                    & 55.91 & \textbf{1.02} & \textbf{0.58} & 3.14  & 56.55 & \textbf{0.72}  & \textbf{0.63}  & \textbf{2.29}  & 41.64 & \textbf{1.01}  & \textbf{0.86}  & 2.99  & \textbf{65.86} & \textbf{1.03} & \textbf{0.93} & \textbf{2.11}  \\ \bottomrule
\end{tabular}}}
\end{table*}

\begin{table}[]
\caption{The calibration results of CTC-based models on the Chinese STR benchmark. The best method is highlighted in bold.}
\label{zh-str-ctc}
\resizebox{\linewidth}{!}{
\setlength{\tabcolsep}{0.85mm}{
\begin{tabular}{l|cccc|cccc}
\toprule
\multirow{2}{*}{Method} & \multicolumn{4}{c|}{CRNN}    & \multicolumn{4}{c}{TRBC}      \\ 
                        & Acc   & ECE  & ACE  & MCE   & Acc   & ECE   & ACE   & MCE   \\ \midrule
CTC                 & \textbf{41.10} & 8.62 & 8.62 & 21.85 & \textbf{58.07} & 15.02 & 15.02 & 39.80 \\PSSR                    & 40.44 & \textbf{0.64} & \textbf{0.48} & \textbf{3.38}  & 57.25 & \textbf{0.79}  & \textbf{0.73}  & \textbf{1.78}  \\ \bottomrule
\end{tabular}}}
\end{table}

\begin{table}[]
\caption{The calibration results of U2-Tfm and U2-CTC on AISHELL-1. The best method is highlighted in bold.}
\label{sr}
\resizebox{\linewidth}{!}{
\setlength{\tabcolsep}{2.6mm}{
\begin{tabular}{l|c|cccc}
\toprule
Models                  & Method       & Acc            & ECE           & ACE           & MCE           \\ \midrule
\multirow{2}{*}{U2-Tfm} & NLL & \textbf{58.81} & 22.75         & 22.75         & 50.85         \\
                        & PSSR         & 57.36          & \textbf{2.21} & \textbf{2.06} & \textbf{7.01} \\ \midrule
\multirow{2}{*}{U2-CTC} & CTC & \textbf{58.14} & 20.20         & 20.20         & 41.28         \\
                        & PSSR         & 57.44          & \textbf{2.47} & \textbf{2.35} & \textbf{3.82} \\
\bottomrule
\end{tabular}}}
\end{table}

\subsection{Comparison with State-of-the-arts}
In this section, we compare the proposed method against the state-of-the-art method on the two tasks: scene text recognition (STR) and speech recognition (SR).

\subsubsection{Results on STR}
We present the quantitative calibration results of attention-based models on the English STR benchmark in Table~\ref{en-str}. The results show that our proposed PSSR outperforms all the compared state-of-the-art methods across all the models in terms of ECE, ACE, and MCE metrics.
Among other comparison methods, the two calibration methods for sequential data, GLS and CASLS, generally perform better than the methods for scalar data and achieve the second-best performances. Compared with the sub-optimal GLS method, particularly in the TRBA model, the proposed method still reduces 0.56\%, 0.62\%, and 3.18\% in ECE, ACE, and MCE, respectively. 
Moreover, Table~\ref{en-str-ctc} reports the calibration results of CTC-based models on the English STR benchmark. Compared with the uncalibrated models trained with CTC loss, the models trained with PSSR perform much better in terms of all the metrics, including accuracy, ECE, ACE, and MCE. 
Combined with the above, the satisfying performance demonstrates that the proposed method can be well adapted to the model with different decoding schemes.

We further verify the effectiveness of our method on the Chinese STR benchmark, and the calibration results of attention and CTC models are presented in Table~\ref{zh-str} and \ref{zh-str-ctc}, respectively.  Notably, our PSSR outperforms other approaches and sets a new state-of-the-art with better accuracy and confidence calibration on almost all the models.

\subsubsection{Results on SR}
Table~\ref{sr} reports the calibration results of uncalibrated models and PSSR on the AISHELL-1 dataset. 
As shown, the proposed PSSR performs better than uncalibrated models in ECE, ACE, and MCE metrics.
Compared to uncalibrated models, the proposed PSSR reduces 20.54\%, 20.69\%, 43.84\% in terms of ECE, ACE, and MCE on the attention-based model and reduces 17.73\%, 17.85\%, 37.46\% in terms of ECE, ACE, and MCE on the CTC-based model.
More results are presented in the appendix.

\begin{table}[!ht]
\centering
\caption{Corrupted calibration results on the English STR benchmark. Uncal is short for Uncalibrated. The best method is highlighted in bold.}
\label{data-shift-table}
\setlength{\tabcolsep}{0.6mm}{
\resizebox{\linewidth}{!}{
\begin{tabular}{c|c|cccc|cccc}
\toprule
\multirow{2}{*}{Corruption}    & \multirow{2}{*}{Method} & \multicolumn{4}{c|}{TRBA}      & \multicolumn{4}{c}{TRBC}      \\ 
                               &                         & Acc   & ECE   & ACE   & MCE   & Acc   & ECE   & ACE   & MCE   \\ \midrule
\multirow{2}{*}{Speckle Noise} & Uncal                & 65.71 & 3.80  & 3.85  & 15.83 & 65.63 & 1.51  & 1.46  & \textbf{7.59}  \\
                               & PSSR                    & \textbf{67.01} & \textbf{0.64}  & \textbf{0.66}  & \textbf{5.84}  & \textbf{66.45} & \textbf{1.19}  & \textbf{0.54}  & 9.26  \\ \midrule
\multirow{2}{*}{Gaussian Blur} & Uncal               & 42.10 & 19.10 & 19.10 & 57.63 & 40.52 & 14.49 & 14.50 & 44.80 \\
                               & PSSR                    & \textbf{42.29} & \textbf{2.45}  & \textbf{2.55}  & \textbf{10.92} & \textbf{40.50} & \textbf{1.45}  & \textbf{1.25}  & \textbf{7.80}  \\ \midrule
\multirow{2}{*}{Spatter}       & Uncal               & 59.91 & 4.12  & 4.12  & 11.79 & 58.12 & 2.23  & 1.89  & 6.41  \\
                               & PSSR                    & \textbf{61.68} & \textbf{1.06}  & \textbf{0.86}  & \textbf{4.89}  & \textbf{58.82} & \textbf{1.99}  & \textbf{1.87}  & \textbf{5.13}  \\ \midrule
\multirow{2}{*}{Saturate}      & Uncal                & 81.04 & 3.95  & 3.95  & 16.96 & 80.41 & 2.56  & 2.48  & 17.56 \\
                               & PSSR                    & \textbf{81.56} & \textbf{0.74}  & \textbf{0.54}  & \textbf{7.07}  & \textbf{80.92} & \textbf{0.64}  & \textbf{0.38}  & \textbf{6.78}  \\ \bottomrule
\end{tabular}
}
}
\end{table}
%%%%%%%%%%%%%%%%%%%%%%%%%%%%%%%%%%%%%%%%%%%%%%%%%%%%%%%%%%%%%%%%%%%%%%%%%
\begin{figure}[b]
\begin{center}
%\fbox{\rule{0pt}{2in} \rule{0.9\linewidth}{0pt}}
\includegraphics[width=1.0\linewidth]{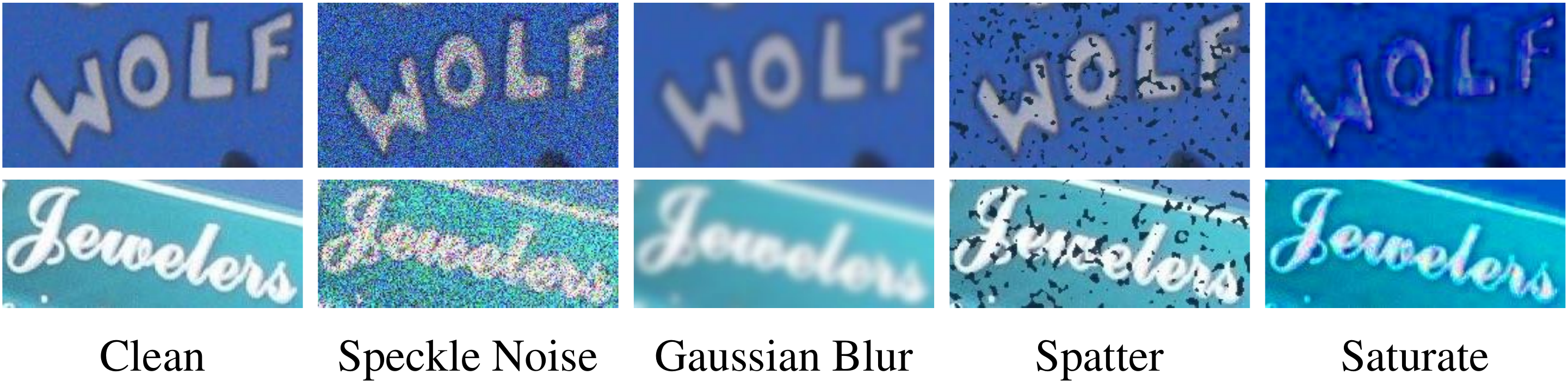}
\end{center}
   \caption{Clean and four corruption examples.}
\label{data-shift}
\end{figure}
%%%%%%%%%%%%%%%%%%%%%%%%%%%%%%%%%%%%%%%%%%%%%%%%%%%%%%%%%%%%%%%%%%%%%%%%%
\subsection{Calibration Performance under Dataset Shift}
The DNNs are discovered to be overconfident and highly uncalibrated under the condition of data shift.
Inspired by~\cite{hendrycks2019benchmarking}, the data distribution drift test datasets are derived from the English benchmark test dataset after four diverse corruption types, including speckle noise, Gaussian blur, spatter, and saturate.
Figure \ref{data-shift} shows the clean and the four corrupted examples.
Table~\ref{data-shift-table} reports the calibration results of uncalibrated models and PSSR on the English STR benchmark of corrupted datasets, which demonstrates that the model trained with the proposed PSSR can still achieve a good calibration performance even under data shift. And compared with the state-of-the-art calibration methods, our method performs best in terms of all the metrics across all the drift datasets. 
And the details on other methods and their corrupted calibration results are presented in the appendix.

%%%%%%%%%%%%%%%%%%%%%%%%%%%%%%%%%%%%%%%%%%%%%%%%%%%%%%%%%%%%%%%%%%%%%%%%%

\begin{figure}
  \centering
  \begin{subfigure}{0.49\linewidth}
    \includegraphics[width=1\linewidth]{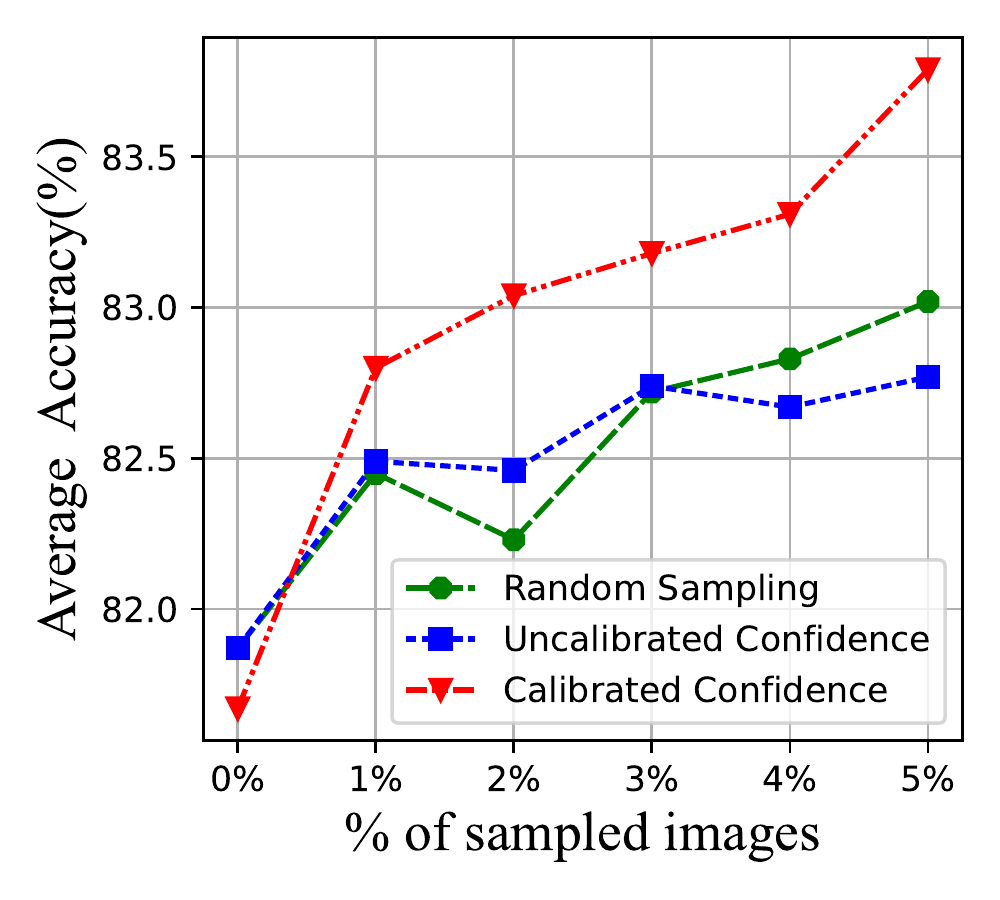}
    \caption{TRBA}
  \end{subfigure}
  \hfill
  \begin{subfigure}{0.49\linewidth}
    \includegraphics[width=1\linewidth]{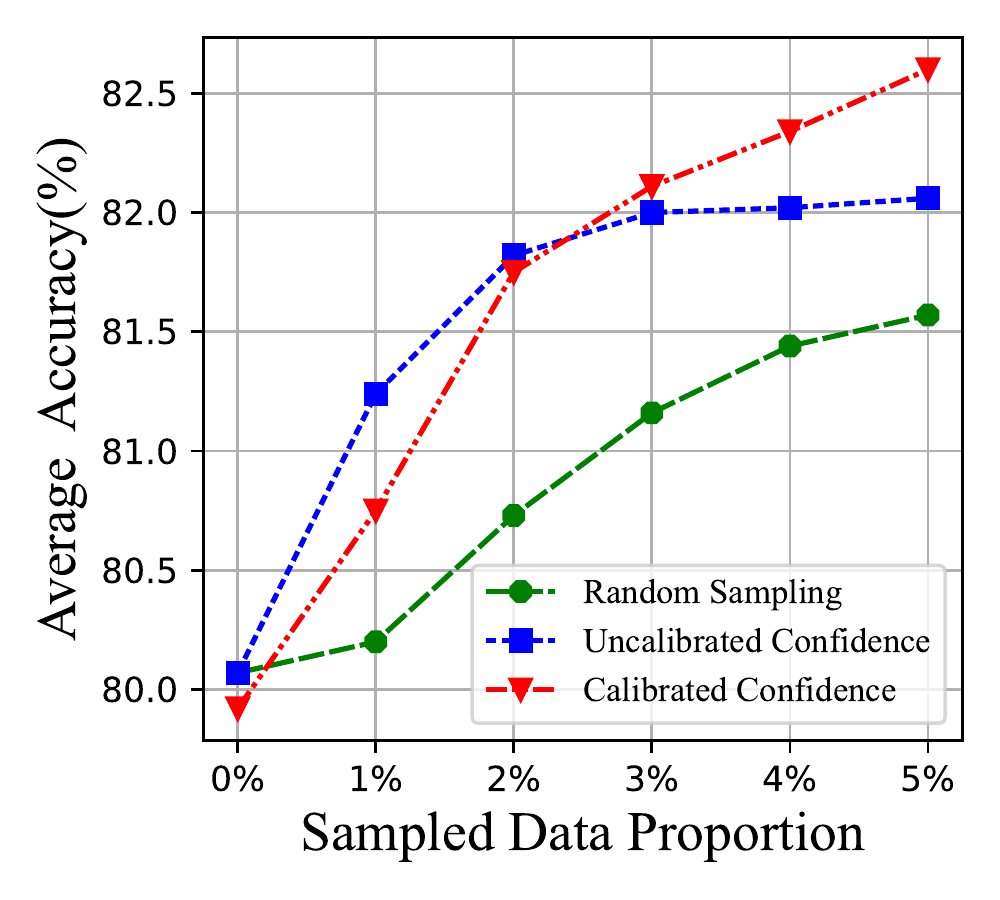}
    \caption{TRBC}
  \end{subfigure}
  \caption{The results of active learning task on TRBA and TRBC.}
  \label{al}
\end{figure}

\subsection{Downstream Application}
We argue that calibration benefits the downstream active learning task when adopting a confidence-based query strategy. 
In general, active learning trains an initial model based on a small amount of labeled data, and then a query strategy is applied to the output of models to select the most informative samples with the least confidence for an oracle to annotate. The model is then retrained with the additional labeled data. The above process will be repeated until model accuracy is satisfied or the labeling resource is exhausted.

The active learning experiment is conducted on the English STR benchmark, where an attention-based TRBA model and a CTC-based TRBC model are adopted. Specifically, only 10\% of training samples are used initially to train the base model. Then, 1\% samples of the unlabeled data pool (consisting of the remaining 90\% of training samples) are queried, combining the original labeled samples to retrain the model.
We compare three query strategies: random sampling, least uncalibrated confidence, and confidence calibrated with our PSSR. And the querying process is iterated five times.

Fig.~\ref{al} shows the average accuracy on the test set against the percentage of images sampled from the unlabeled data pool for different models. It can be seen that the accuracy using the confidence-based strategy performs better than other query strategies. And it further outperforms the uncalibrated confidence-based strategy with accuracy improvement by 1.02\% and 1.03\% after the final iteration on TRBA and CRNN, respectively.

%%%%%%%%%%%%%%%%%%%%%%%%%%%%%%%%ly%%%%%%%%%%%%%%%%%%%%%%%%%%%%%%%%%%%
\section{Conclusion}
Despite the superior performance of deep sequence recognition models, they have been proven to suffer from the over-confidence dilemma. In this paper, we investigate the overconfidence problem of the DSR model and discover that tokens/sequences with higher perception and semantic correlations to the target ones contain more sufficient and correlated information to supervise the regularization of labels and facilitate more effective regularization. Motivated by the observation, we propose a Perception and Semantic aware Sequence Regularization framework, which explores perceptively and semantically correlated tokens/sequences as regularization. Comprehensive experiments are conducted on classic DSR tasks: scene text and speech recognition, and our method achieves state-of-the-art confidence calibration performance.
In the future, we will explore more effective strategies to conjointly utilize perception and semantic information for better DSR model calibration.

\vspace{-.35em}
\section*{Acknowledgment}
\vspace{-.35em}
This research was supported in part by NSFC (Grant No. 62176093, 61673182, 62206060), Key Realm R\&D Program of Guangzhou (No. 202206030001), Guangdong Basic and Applied Basic Research Foundation (No. 2021A1515012282) and Guangdong Provincial Key Laboratory of Human Digital Twin (No. 2022B1212010004).

%%%%%%%%% REFERENCES
{\small
\bibliographystyle{ieee_fullname}
\bibliography{manuscript}
}

\end{document}